\documentclass[twoside]{article}
\usepackage{amsthm,amsmath}
\usepackage{amssymb}
\usepackage{bm}
\usepackage{bbm}
\usepackage{url}
\usepackage{url}
\usepackage{graphicx}
\usepackage{color}
\usepackage{nicefrac}
\usepackage{tikz}
\usepackage{graphicx}
\usepackage{caption}
\usepackage{float}
\usepackage[makeroom]{cancel}

\usepackage[accepted]{aistats2015}
\usepackage[aboveskip=0pt]{caption}
 
\definecolor{purple}{RGB}{191,85,236}

\begin{document}

\twocolumn[

\aistatstitle{Criticality \& Deep Learning I: Generally Weighted Nets}

\aistatsauthor{ Dan Oprisa\\ \texttt{dan.oprisa@critical.ai} \And Peter Toth\\ \texttt{peter.toth@critical.ai}}

\aistatsaddress{ \\CriticalAI \\ http://www.critical.ai } 
]

\begin{abstract}

Motivated by the idea that criticality and universality of phase transitions might play a crucial role in achieving and sustaining learning and intelligent behaviour in biological and artificial networks, we analyse a theoretical and a pragmatic experimental set up for critical phenomena in deep learning. On the theoretical side, we use results from statistical physics to carry out critical point calculations in feed-forward/fully connected networks, while on the experimental side we set out to find traces of criticality in deep neural networks. This is our first step in a series of upcoming investigations to map out the relationship between criticality and learning in deep networks. 

\end{abstract}

\section{Introduction}

Various systems in nature display patterns, forms, attractors and recurrent behavior, which are not caused by a law per se; the ubiquity of such systems and similar statistical properties of their exhibit order has lead to the term "universality", since such phenomena show up in cosmology, the fur of animals \cite{per_bak_nature}, chemical and physical systems \cite{Scale_Invariance_Lesne_Lagues}, landscapes, biological prey-predator systems and endless many others \cite{pruessner}. Furthermore, because of universality, it turns out that the most simplistic mathematical models exhibit the same statistical properties when their parameters are tuned correctly. As such it suffices to study N-particle systems with simple, "atomistic" components and interactions since they already exhibit many non-trivial emergent properties in the large N limit. Certain "order" parameters change behavior in a non-classical fashion, for specific noise levels. Using the rich and deep knowledge gained in statistical physics about those systems, we map the mathematical properties and learn about novel behaviors in deep learning set ups. Specifically we look at a collection of N units on a lattice with various pair interactions; when the units are binary spins with values ($\pm 1$), the model is known as a Curie-Weiss model. From a physical point of view, this is one of the basic, analytically solvable models, which still possesses the rich emergent properties of critical phenomena. However, given its general mathematical structure, the model has already been used to explain  population dynamics in biology \cite{bio_sys_poised_crit}, opinion formation in society \cite{Cover_Joy_Elements_of_Information_Theory}, machine learning \cite{fukushima_self_org, barra_phase_boltzmann, hopf_boltzmann} and many others \cite{hopfield_nn_phys_sys, peterson_anderson_mft}. All those systems, with a rich and diverse origination, posses almost identical behavior at criticality. In the latter case of machine learning, the Curie-Weiss model encodes information about fully connected and feed-forward architectures to first order.
Similar work was done in \cite{loss_surface, spinglass_bialek}, where insights from Ising models and fully connected layers are drawn and applied to net architectures; in \cite{energyae} a natural link between the energy function and an autoencoder is established.
We will address the generalisation of fully connected system and understand its properties, before moving to the deep learning network and applying there the same techniques and intuition.

The article is organised as follows: section \ref{sect:crit_phenomen} gives a short introduction of critical systems and appropriate examples from physics; in section \ref{sect:crit_dl_nets} we map a concrete, non-linear, feed forward net to its physical counterpart and discuss other architectures as well; then we turn to investigating the practical question whether we can spot traces of criticality in current deep learning nets in \ref{sect:exp_results}. Finally we summarise our findings in \ref{sect:outlook} and hint at future directions for the rich map between statistical systems and deep learning.

\section{Brief summary of critical phenomena}\label{sect:crit_phenomen}

Critical phenomena were first thoroughly explained and analysed in the field of statistical mechanics, although they were observed in various other systems, but lacking a theoretical understanding. The study of criticality belongs to statistical physics and is
an incredibly rich and wide field, hence we can only briefly summarise some few results of interest for the present article; definitely a much more comprehensive coverage can be found, see e.g.\cite{complex_crit, life_order, renorm_stat_phys, origin_order}. In a nutshell, the subject is concerned with the behavior of systems in the neighbourhood of their critical points, \cite{scale_renorm_univers}. One thus looks at systems composed of (families of) many, identical particles, trying to derive properties for macroscopic parameters, such as density or polarisation from the microscopic properties and interactions of the particles; statistical mechanics can hence be understood as a bridge between macroscopic phenomenology (e.g. thermodynamics) and microscopic dynamics (e.g. molecular or quantum-mechanical interacting collections of particles). In a nutshell, criticality is achieved when macroscopic parameters show anomalous, divergent behavior at a phase transition. Depending on the system at hand, the parameters might be magnetisation, polarisation, correlation, density, etc. Specifically it is the correlation function of the "components" which then displays divergent behavior, and signals strong coordinated group behavior over a wide range of magnitudes. Usually it is the noise (temperature) which at certain values will induce the phase transition accompanied by the critical anomalous behavior. Given its relevance in physics and also its mathematical analogy to our deep learning networks, we will briefly review here the Curie-Weiss model with non-constant coupling and examine its behavior at criticality.

\subsection{Curie-Weiss model}

A simplistic, fully solvable model for a magnet is the Curie-Weiss model (CW), \cite{salinas}. It possesses many interesting features, exhibits critical behavior and correctly predicts some of the experimental findings. As its mathematics is later on used in our deep learning setup, we will briefly present main properties and solutions for the sake of self-consistency.

The Hamiltonian of the CW model is given by

\begin{equation}
    \centering
    H =  -\frac{J}{2N}\sum_{ij}^N s_is_j 
    - b \sum_{i}^N s_i
\end{equation}

Here the $s_i$ are a collection of interacting "particles", in our physical case, spins, that interact with each other via the coupling $J$; they take values $(\pm 1)$ and interact pairwise with each other, at long distances; the inclusion of a factor of $\frac{1}{N}$ multiplying the quadratic spin term makes this long-range interaction tractable in the large $N$ limit. Furthermore, there is a directed external magnetic field which couples to every spin via $b$. Since the coupling between spins is a constant and since every spin interacts with every other spin (except self-interactions, which is accounted by a factor of $\frac{1}{2}$) the Hamiltonian can be rewritten to

\begin{equation}
    \centering
    H =  -\frac{J}{2N}\bigg(\sum_{i}^N s_i\bigg)^2
    - b \sum_{i}^N s_i
\end{equation}

With $\beta = \nicefrac{1}{kT}$ being the inverse temperature the partition function can be formulated

\begin{align}
Z &= \sum_{s_i\in\{\pm 1\}}e^{-\beta H(s)}\\
 &= \sum_{s_i\in\{\pm 1\}} \exp \beta\bigg[   \frac{J}{2 N}\bigg(\textstyle\sum_{i}^N
 s_i\bigg)^2    + b \textstyle\sum_{i}^N s_i\bigg]
\end{align}

which can be fully solved, \cite{salinas}, summing over each of the $2^N$ states; given an explicit partition $Z$, the free energy can be computed via 

\begin{align}
F = - kT \ln Z 
\end{align}

Once we have $F$ various macroscopic values of interest can be inferred such as the magnetisation of the system, aka first derivative of $F$ wrt. $b$. This is a so called "order parameter", which carries various other denominations, such as polarisation, density, opinion imbalance, etc. depending on the system at hand. It basically measures how arranged or homogeneous the system is under the influence of the outside field which couples to the spins via $b$. A full treatment and derivation of the model including all its critical behavior can be found in \cite{cw_full_solution}, from where we get the equation of state for the magnetisation

\begin{align}
\label{magnetisation}
m = b\tanh \big( \frac{K}{b}m + \frac{b}{T} \big) 
\end{align}

with $K = (\frac{J}{T})^{1/2}$. The analysis of this equation for various temperatures $T$ and couplings $J$, $b$ reveals a phase transition at the critical temperature $T_c = J$. Introducing the dimensionless parameter $t=(\nicefrac{T}{T_c}-1)$ and expanding (\ref{magnetisation}) in small couplings the famous power law dependence on temperature for the magnetisation emerges:

\begin{align}
\label{mag_cw}
\centering
m \simeq 
\sqrt{3}\frac{ (K-1)^{1/2} }{K^{3/2}} &\sim |t|^{1/2 }
\end{align}

Here we recognise one of the very typical power laws which are ubiquitous to critical systems.
The quantity we are most interested in though is the second derivative of the free energy $F$ wrt. $b$, which is basically the 2-point correlation function of the spins $s_i$. Again, expanding the second derivative of the free energy in small couplings and looking in the neighbourhood of the critical temperature $T_c$ yields 

\begin{equation}
\label{crit_corr}
\centering
\langle s_i,s_j\rangle \sim \frac{b^2}{T_c}|t|^{-1}
\end{equation}

again displaying power law behavior with a power coefficient $\gamma = 1$. The innocent looking equation \ref{crit_corr} has actually tremendous consequences, as it implies that correlation does not simply restrict to nearest neighbours but goes over very long distances only slowly decaying; further, because of the power law behavior, there will be self-similar, fractal patterns in the system: islands of equal magnetisation will form within other islands and so on, all the way through all scales.
Also, the correlation diverges at the criticality point $T_c$. We will carry out the explicit calculations for our case of interest - non-constant matrix couplings - later one, in section \ref{sect:real_nets}.

\subsection{Criticality in real-world networks}

Two of the main motivations why we look for criticality and exploit on it in artificial networks, are the universal arising of this phenomenon as well as various hints of its occurrence in biological \cite{bio_sys_poised_crit} and neural systems \cite{crit_brain_dynam, weak_pair_wise_corr_popul}; once systems get "sizable" enough, gaining complexity, critical behavior emerges, which also applies to man-made nets \cite{stat_mech_complex_nets}. Various measures can be formulated to detect criticality, and they all show power law distribution behavior. In the world wide web, e.g. the number of links to a source, and the number of links away from a source, both exhibit power law distribution

\begin{equation}
\centering
P(k) \sim k^{-\gamma}
\end{equation}

for some power coefficient $\gamma \neq 0$. Similar behavior can be uncovered in various other networks, if sizable enough, such as citation behavior of scientific articles, social networks, etc. A simple, generic metric to detect criticality in networks is the degree distribution, defined as the number of (weighted) links connecting to one node. 

Further, also the correlation between nodes is non-trivial, such that nodes with similar degree have higher probability of being connected than nodes of different degree \cite{stat_mech_complex_nets}, chapter VII. We will follow a similar path as proposed above and grow an experimental network with new nodes having the simplest preferential and directed attachment towards existing nodes, as a function of their degree:

\begin{equation}
\centering
\Pi(k) \sim k^{\alpha}
\end{equation}

Here, $\Pi(k)$ denotes the probability that some node will grow a link to another node of degree $k$. Hence, every new node, will {\itshape prefer} nodes with higher degrees, leading to the overall power distribution observed in the real world systems.
Additional metrics we look at are single neuron activity as well as layer activity and pattern behavior; more details on that in section \ref{sect:exp_results}. 

\section{Criticality in deep learning nets}
\label{sect:crit_dl_nets}

\subsection{From feed-forward to fully connected architecture}
\label{sect:real_nets}

We will focus now on a feed-forward network, with two layers, $a_i$ and $b_j$ connected via a weight matrix $w_{ij}$; In order to probe our system for criticality, we write down its Hamiltonian

\begin{equation}
    \centering
    H =  -\frac{1}{2N}\sum_{ij}^N w_{ij}a_ib_j 
    - h \sum_{i}^N b_i
\end{equation}

which has been first formulated in the seminal paper \cite{hopfield_nn_phys_sys}. Here, the values of the $a$ and $b$
are $\{0,1\}$. Further, by absorbing the biases $b_i$ in the usual way we can assume our weight matrix has the form: 

\begin{equation}
\centering
\label{w_plus_h}
W = \begin{pmatrix}
2Nh & 0 & \cdots & 0 \\
2Nh & w_{11} & \cdots & w_{1n} \\
\vdots  & \vdots  & \ddots & \vdots  \\
-2Nh & w_{n1} & \cdots & w_{nn} 
\end{pmatrix}
\end{equation}

while the $V_i$ read $(1,V_1,\cdots, V_N)$.

This Hamiltonian describes a two layer net containing rectified linear units (ReLU) in the $b$-layer with a common bias term $h$. The weight matrix $w_{ij}$ sums the binary inputs coming from the $a_i$ and those are fed into $b_i$; depending whether the ReLU threshold has been reached, $a_i$ is activated, hence the binary values allowed for both, inputs and $b$-layer.

Further, we show in appendix \ref{app1}, that the partition function is up to a constant the same for the units taking values in $\{\pm 1\}$ or $\{0,1\}$. By redefining $N+1 \rightarrow N$ We can then formulate the partition function as

\begin{equation}
\label{part_fc}
    \centering
    Z =  \sum_{a,b\in\{\pm 1\}}  e^{-\frac{\beta}{2N}\sum_{ij} W_{ij}a_ib_j }
\end{equation}

where $\beta$ is the inverse temperature $\nicefrac{1}{T}$. This is the partition function of a bipartite graph with non-constant connection matrix $w$.

However, it turns out, that the partition function of the fully connected layer is the highest contribution (1st order) of our feed forward network (see appendix \ref{app2}), hence further simplifying the expression to

\begin{equation}
\label{part_fc}
    \centering
    Z =  \sum_{s_i\in\{\pm 1\}}  e^{-\frac{\beta}{2N} \sum_{ij} W_{ij}s_is_j }
\end{equation}

We will now proceed and compute the free energy $F$, defined as $F = -T\ln{Z}$, using the procedure presented in \cite{peterson_anderson_mft}. From the free energy we
then find all quantities of interest, especially the 2-point correlation function of the neurons.

\subsection{Fully connected architecture with non-constant weights}

In order to solve the CW model analytically, one has to perform the sum over spins, which is hindered by the quadratic term $s_is_j$. The standard way to overcome this problem is the gaussian linearisation trick which replaces quadratic term by its square root - linear in $s_i$  and one additional continuous variable - the "mean" field, which is being integrated over entire $\mathbbm{R}$:

\begin{equation}
e^{a^2} = \frac{1}{\sqrt{2\pi}}\int_{-\infty}^\infty dx e^{-x^2/2 + \sqrt{2}ax}
\end{equation}

which in physics, is known as the Hubbard–Stratonovich transform.

Unfortunately our coupling is not scalar, and hence we will linearise the sum term by term to keep track of all the weight matrix entries.
First we will insert $N$ identities via the Dirac delta function into our Hamiltonian as used in  (\ref{part_fc}):

\begin{align}
\centering
H(s) &=  -\frac{1}{2N}\sum_{ij}^N s_iW_{ij}s_j \\
&= -\frac{1}{2N} \prod_k\int_{-\infty}^\infty dV_k\delta(s_k-V_k)\sum_{ij}^N V_iW_{ij}V_j \nonumber\\
&= \prod_k\int_{-\infty}^\infty dV_k\delta(s_k-V_k) H(V) \nonumber
\end{align}

With the definition of the delta function 
$\delta(x) = \frac{1}{2\pi i}\int_{-i\infty}^{i\infty}dye^{xy}$
the partition function (\ref{part_fc}) reads now

\begin{align}
&Z(s) =  \prod_k\int_{-\infty}^\infty dV_k\delta(s_k-V_k)
\sum_{s_i\in\{\pm 1\}}e^{-\beta H(V)}\\
&\sim  \prod_k\int_{-\infty}^\infty dV_k\int_{-i\infty}^{i\infty}dU_k\sum_{s_i\in\{\pm 1\}} e^{U_k(s_k-V_k)}
e^{-\beta H(V)}\nonumber\\
&=  \prod_k\int_{-\infty}^\infty dV_k\int_{-i\infty}^{i\infty}dU_k e^{-U_kV_k + \ln(\cosh U_k)}
e^{-\beta H(V)}\nonumber
\end{align}

As already stated, we could perform the sum over the binary units $s_i$, since they show up linearly in the exponential after the change of variables via delta identity\footnote{In general we're not interested in numerical multiplicative constants, as later on, when logging the partition and computing the free energy, those terms will be simple additive constants without any contribution after differentiating the expression}; we effectively converted the sum over binary values $\{\pm 1\}$ into integrals over $\mathbbm{R}$, leading to

\begin{equation}
\label{Z_mft}
    \centering
    Z =  c\,\prod_i \int_{-\infty}^{\infty}dV_i \int_{-i\infty}^{i\infty}dU_i \, e^{-H^{g}(V,U,T)}
\end{equation}

with a generalised Hamiltonian

\begin{align}
\label{gen_ham}
H^{g} =  &-\frac{\beta}{2N}\sum_{ij} W_{ij}V_iV_j +\sum_i\big[ U_iV_i - \ln{(\cosh{U_i})} \big]\nonumber\\    = &-\frac{\beta}{2N}\textstyle\sum_{ij} w_{ij}V_iV_j - \beta h\sum_iV_i\nonumber\\ 
&+\sum_i\big[ U_iV_i -\ln{(\cosh{U_i})} \big]
\end{align}

Ultimately we are interested in the free energy per unit, which contains the partition function, via

\begin{equation}
\label{helmholtz}
\centering
F = \lim_{N\to\infty} (-T\ln{Z})/N
\end{equation}

From $F$ we can now obtain all quantities of interest via derivatives, in our case with respect to $h$. The partition function $Z$ still contains a product of double integrals, which can be solved via the saddle point approximation; we recall here the one-dimensional case 

\begin{equation}
\label{saddle_eq}
\int_{-\infty}^\infty dx e^{-f(x)} \approx \big(\frac{2\pi}{f''(x_0)}\big)^{1/2}e^{-f(x_0)}
\end{equation}

where $x_0$ is the stationary value of $f$ and $f''(x_0)$ is in our case the Hessian evaluated at the stationary point:

\begin{align}
H^{g}_{V_iV_j} &=  -\frac{\beta}{N}w_{ij}\\
H^{g}_{U_iU_j} &=  -\delta_{ij}(1-\tanh^2U_i)\nonumber\\
H^{g}_{V_iU_j} &=  \delta_{ij}\nonumber\\
\end{align}

while $H^{g}$ is given in (\ref{gen_ham}).

The expression \ref{Z_mft} can now be computed by applying simultaneously the saddle point conditions for both integrals. The stationarity conditions\footnote{We keep in mind that we enlarged $W$ to contain $h$ as well, hence the explicit equations are $h$ dependent} for $V_i$ and $U_i$ give

\begin{align}
\label{stationarity}
\frac{\partial H^{g}}{\partial V_i} & = -\frac{\beta}{N}\sum_{j}W_{ij}V_j + U_i=  0\\
& = -\beta(\textstyle\sum_{j}w_{ij}V_j/N + h) + U_i=  0\nonumber\\
\frac{\partial H^{g}}{\partial U_i} & = V_i - \tanh{U_i} = 0\nonumber
\end{align}

which combined deliver the self consistency mean field equation of the fully connected layer (\ref{sum_mf}). Further, denoting $H^{g}_0$ the the Hamiltonian satisfying the stationarity conditions, it reads

\begin{align}
\label{saddle_ham}
H^{g}_0 = &\frac{\beta}{2N}\textstyle\sum_{ij} w_{ij}V_iV_j \\
-&\sum_i\ln{\cosh \beta(\textstyle\sum_{j}w_{ij}V_j/N + h) } \nonumber
\end{align}

Equation (\ref{saddle_ham}) already displays manifestly the consistency equation for the mean field, as taking the first derivative wrt. $V_i$ leaves exactly the consistency equation over per its construction;

Now we can rewrite the free energy (\ref{helmholtz}) as

\begin{align}
\centering
F &= \lim_{N\to\infty}\frac{T}{N}\big[H^{g}_0 + \ln\det H^{g}_{hh}\big]\sim \lim_{N\to\infty}\\ 
  &\big[\frac{1}{2N^2}\sum_{ij} w_{ij}V_iV_j
+\frac{1}{N^2}\sum_{ij}\ln[ w_{ij}(1-V^2_i)-1] \nonumber\\
&-  \frac{T}{N}\sum_i\ln\cosh\beta(\textstyle\sum_{j}w_{ij}V_j/N + h)\big]
\nonumber
\end{align}

We need to address now the large $N$ limit; obviously
the second term coming from the determinant clearly vanishes in the large-$N$ limit, as the logarithm is slowly increasing, while we divide through $N^2$;
the first term - a double sum over $V_i$ is of order $N^2$ and hence a well defined average in the limit;
the last term - $\ln\cosh$, when expanded, is again linear in the sum\footnote{The interior sum over $j$ is an  average, hence well defined in the limit; after expansion, we're left with the outer sum (over $i$), which is again a well defined average when divided by $N$ }, and hence a well defined average after dividing through $N$, hence we're left with the free energy

\begin{align}
F =& \frac{T}{N}H_0^{g}\\
=&\frac{1}{2N^2}\sum_{ij} w_{ij}V_iV_j\nonumber\\
&-  \frac{T}{N}\sum_i\ln\cosh\beta(\textstyle\sum_{j}w_{ij}V_j/N + h)\big]\nonumber
\end{align}

We're at the point now, where all quantities of interest can be derived from the free energy $F$; the order parameter (aka magnetisation when dealing with spins) per unit is defined as 

\begin{align}
m &\equiv\frac{dF}{dh} = \left.\frac{\partial F}{\partial h}\right|_{V^{st}} + \cancelto{0}{\frac{\partial F}{\partial V_i}}\left.\frac{\partial V_i}{\partial h}\right|_{V^{st}} 
\end{align}

The second term on the right vanishes identically, as we recognize it being evaluated at the stationarity condition $V^{st}$ for the Hamiltonian. The contribution of the first term is:

\begin{align}
\label{sum_mf}
\sum_{i}w_{ik}V_k/N &= \frac{1}{N}\sum_i\tanh\beta
(\textstyle\sum_{k}w_{ik}V_k/N + h)w_{ik}\nonumber\\
&\Updownarrow\\
V_i &= \tanh\beta
(\textstyle\sum_{k}w_{ik}V_k/N + h)\nonumber
\end{align}

which is (the weighted sum version of) the iconic self-consistency mean field equation of the CW magnet (\ref{magnetisation}).

The critical point, $P_c$ is located where the correlation function diverges for $h\to 0$; the 2-point correlation function (aka susceptibility when dealing with spins) is the second derivative of F, i.e. the derivative of (\ref{sum_mf}) wrt. $h$:

\begin{align}
\label{c_p}
P_c &\equiv \frac{d^2F}{dh^2} = \frac{dm}{dh}\\
&\Updownarrow\nonumber\\
\frac{\partial V_i}{\partial h} &= \beta(
1-V_i^2)(1+ \sum_kw_{ik}\frac{\partial V_k}{\partial h}/N)\nonumber
\end{align}

where we used the original equation (\ref{sum_mf}) for taking the derivatives. 
It is worth contemplating first equations (\ref{sum_mf}) and (\ref{c_p}). They both capture the essence of the criticality of our system, including it's power law behavior. When the weight matrix reduces to a scalar coupling, both equations reduce to the classical CW system and display the behavior shown in (\ref{mag_cw}) and (\ref{crit_corr}). Furthermore, eq. (\ref{c_p}) encodes all the information needed for finding the critical point of matrix system at hand; we recall that all $V$s (and their derivatives) are already implicitly "solved" in terms of $h$ and $w_{ij}$ via the stationarity equation (\ref{sum_mf}) and hence the $V_i$
are just place holders for functions of $w$ and $h$; 
we're thus left with a non-linear system of first order differential equations in $N$ variables, which
will produce poles for specific values of the couplings and temperature at criticality.  

\clearpage

\section{Experimental results}
\label{sect:exp_results}

\begin{figure}[t!]
\centering
\includegraphics[width=6cm]{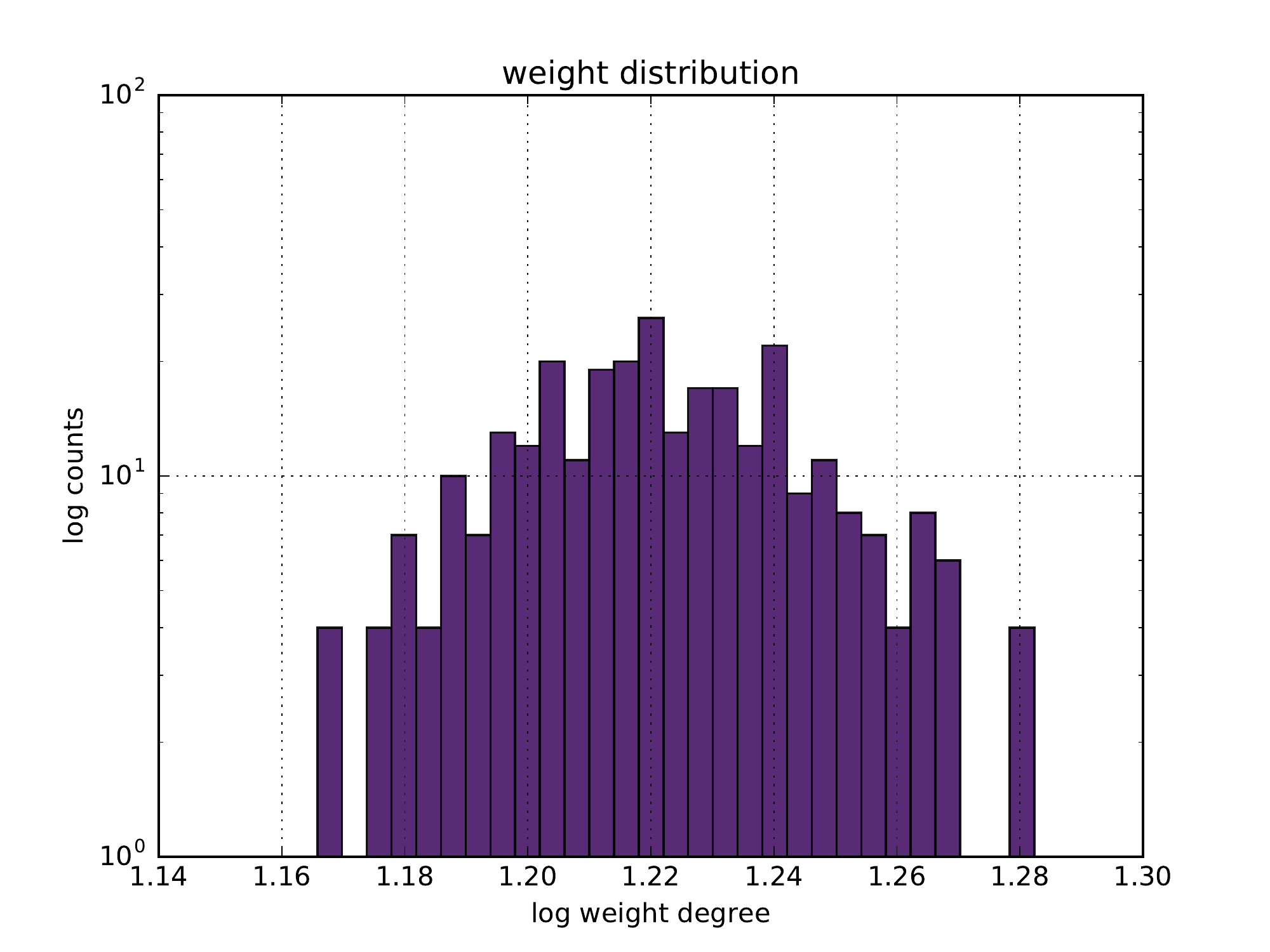}
\caption{Feed-forward net: Layer 3 weight distribution}
\label{fig:ff_weight_W_3_500_400_300_200_200_weight_plot}
\end{figure}

After investigating criticality through the partition function in our theoretical setup, now we turn to a practical question: do current deep learning networks exhibit critical behaviour, or put it differently, can we spot traces of critical phenomena in them? Instead of directly attacking the partition function of real world deep neural nets, we start with the practical observation, that systems at around criticality show off power law distributions in certain internal attributes. 

\begin{figure}[h!]
\centering
\includegraphics[width=6cm]{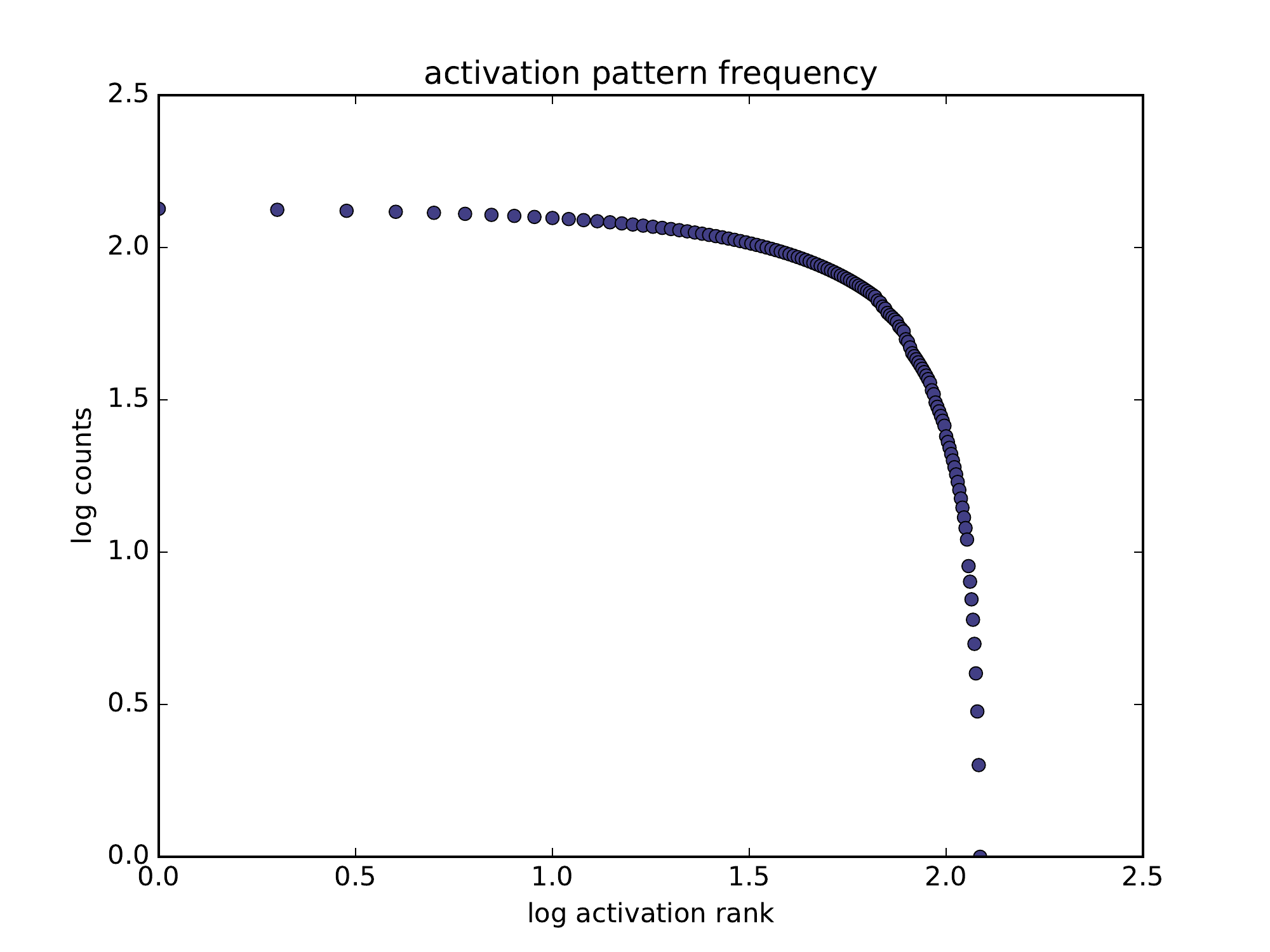}
\caption{Feed-forward net: Log-log plot of layer activation pattern frequencies by rank}
\label{fig:ff_act_500_400_300_200_200_act_plot_layer_2}

\includegraphics[width=6cm]{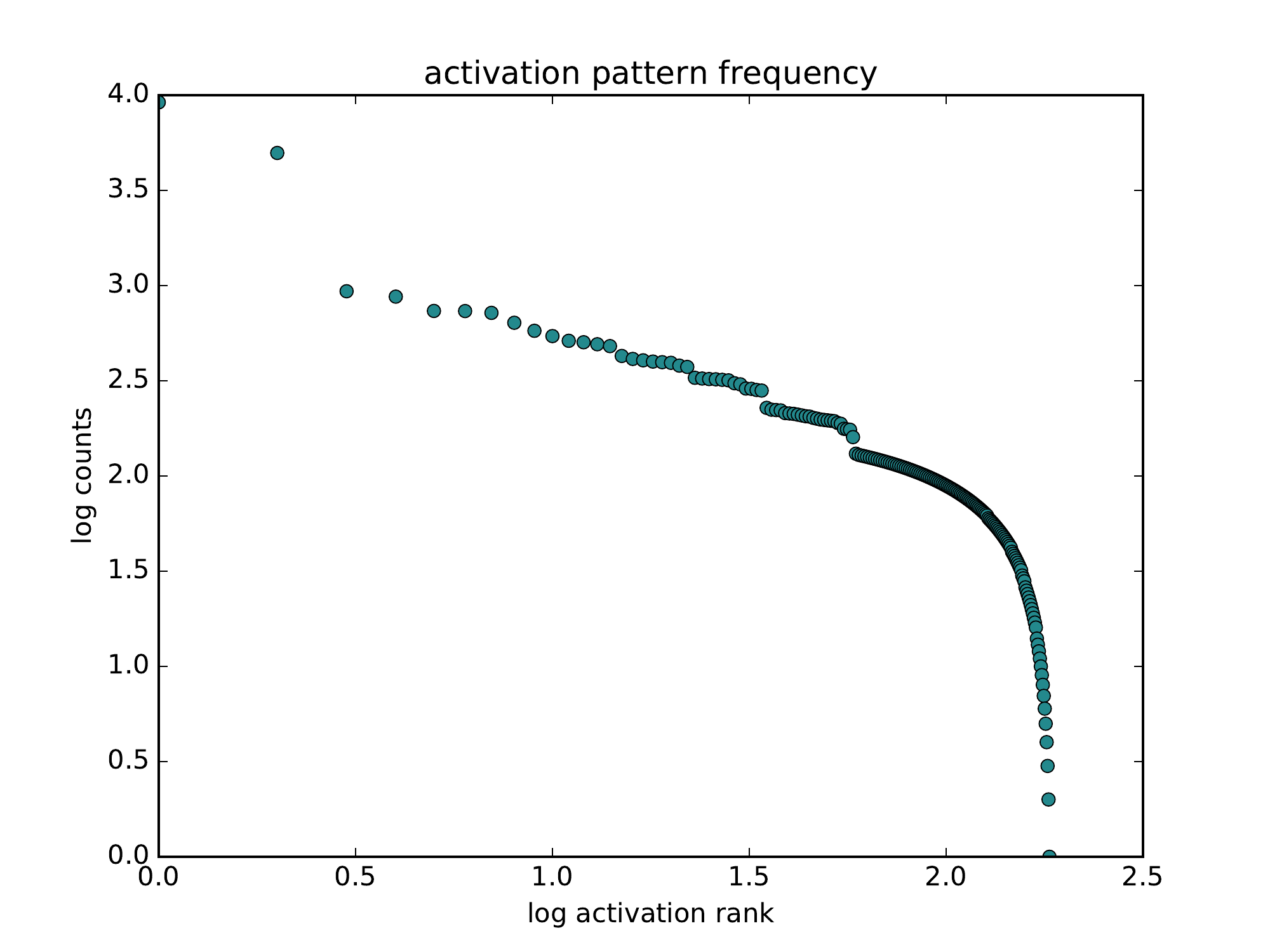}
\caption{Autoencoder: Log-log plot of layer activation pattern frequencies by rank}
\label{fig:ae_act_500_act_plot_layer_1}
\end{figure}

Concretely for networks \cite{stat_mech_complex_nets,zipf_crit_no_fine} we look for traces of power laws in weight distributions, layer activation pattern frequencies, single node activation frequencies and average layer activations.
In the following we will present experimental results for multilayer feed-forward networks, convolutional neural nets and autoencoders.

For all networks we ran experiments on the CIFAR-10 dataset, training each models for 200 epochs using ReLU activations and Adam Optimizer without gradient clipping and run inferences for 100 epochs. The feed forward network had 3 layers with 500, 400 and 200 nodes, the CNN had 3 convolutional layers followed by 3 fully connected layers and the autoencoder had one layer with 500 nodes. 

For weight distributions we looked at sums of absolute values of the outgoing weights at each node, as a weighted order of the node. In fig. \ref{fig:ff_weight_W_3_500_400_300_200_200_weight_plot} we have a log-log plot of counts versus the node order as defined above, and detect no linear behavior.

\begin{figure}[h!]
\centering
\includegraphics[width=6cm]{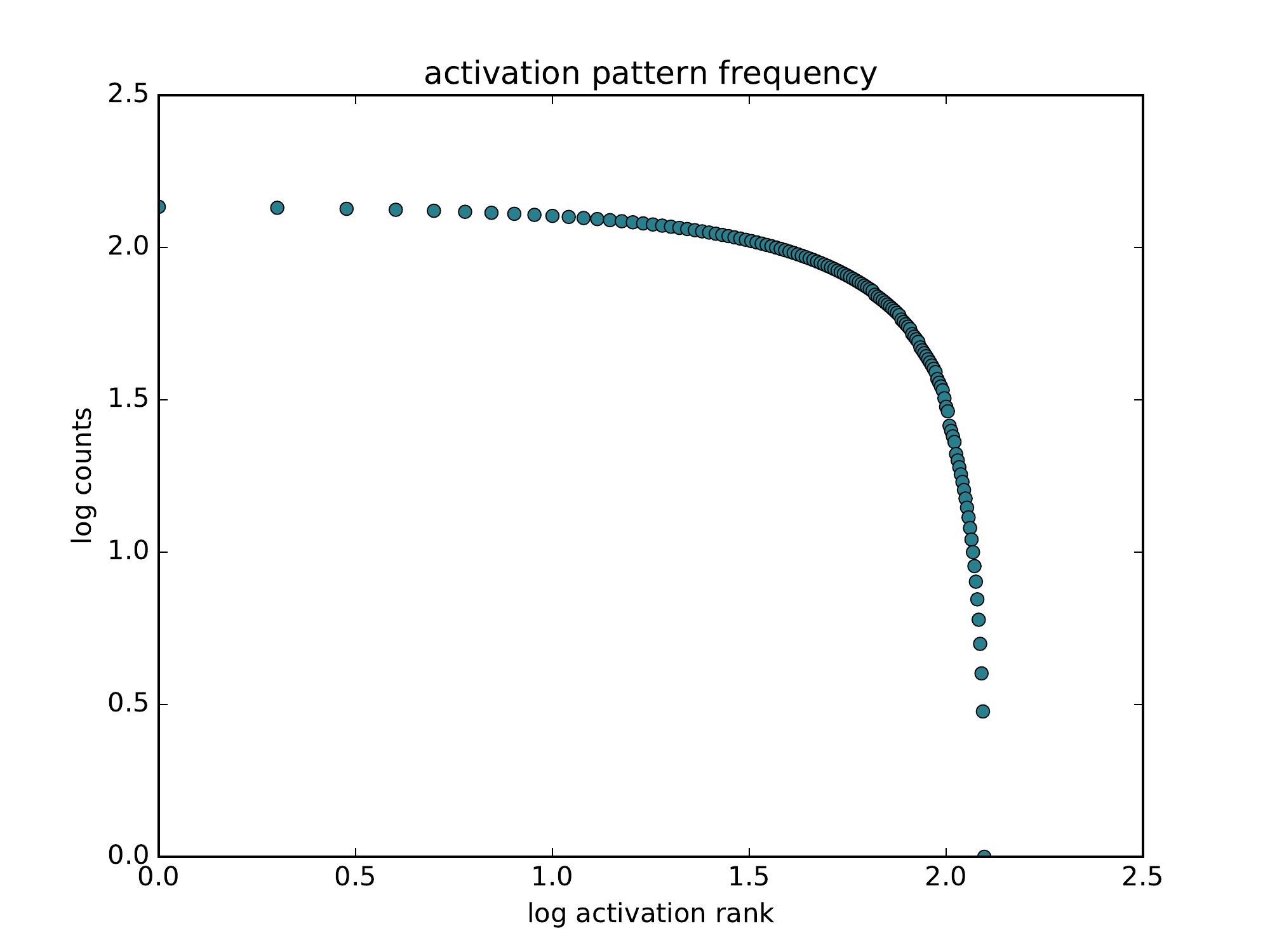}
\caption{CNN: Log-log plot of layer activation pattern frequencies by rank}
\label{fig:cnn_act_conv3_local3_act_plot_layer_3}

\includegraphics[width=6cm]{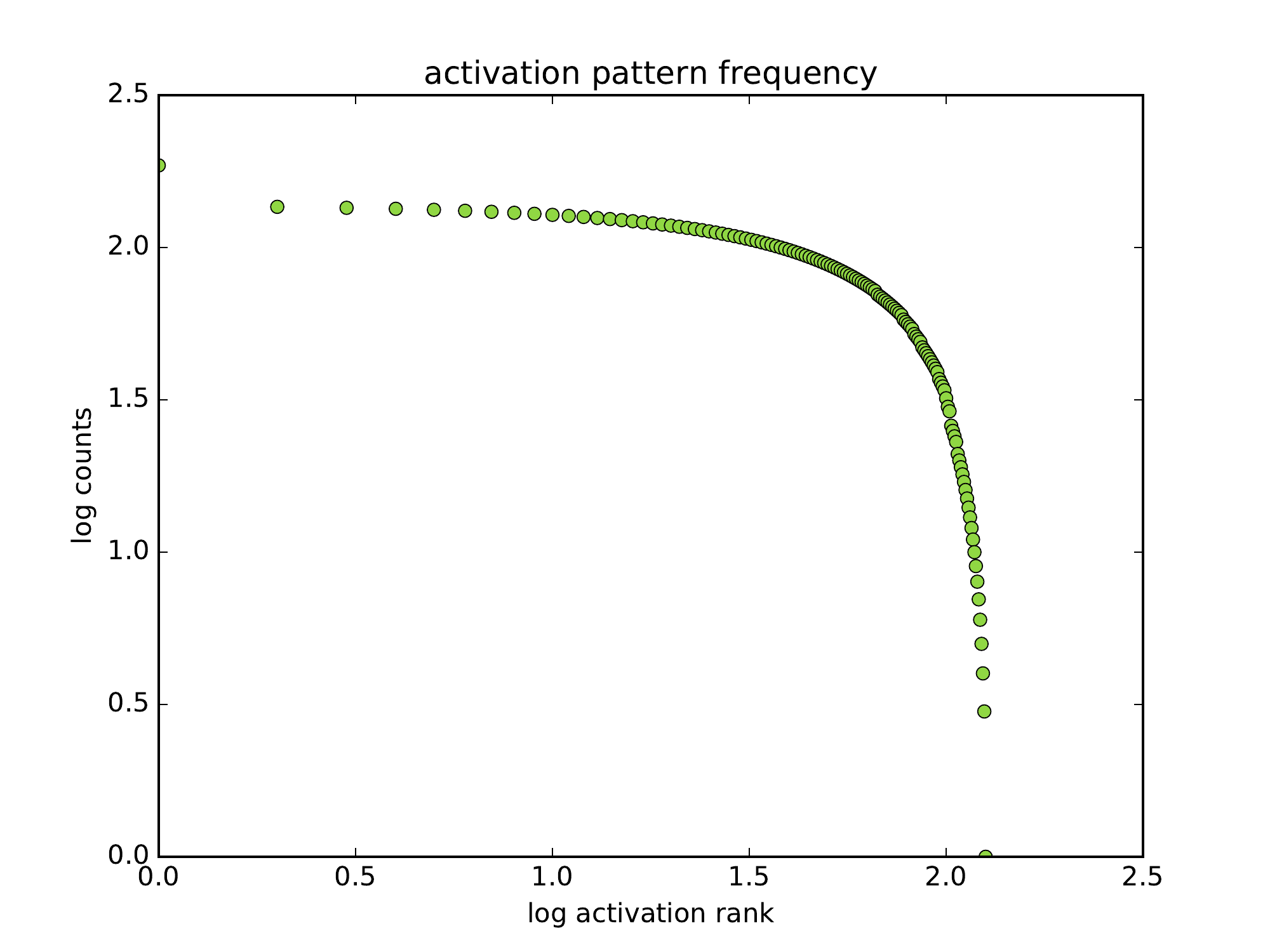}
\caption{CNN: Log-log plot of layer activation pattern frequencies by rank}
\label{fig:cnn_act_conv3_local3_act_plot_layer_4}
\end{figure}

For layer activation patterns we counted the frequency of each layer activations through the inference epochs. Figures \ref{fig:ff_act_500_400_300_200_200_act_plot_layer_2} and \ref{fig:ae_act_500_act_plot_layer_1} are log-log plots of layer activation frequencies versus their respective counts for the feed-forward layer the autoencoder. As we see, the hidden layer activation pattern frequencies of the Autoencoder resembles a truncated straight line, indicating that learning hidden features in unsupervised manner can give rise for scale free, power law phenomena in accordance with the findings of \cite{zipf_crit_no_fine}, but no other architectures show traces of any power law. 

For single node activation frequencies we counted the frequency of each node activations through the inference epochs. 

\begin{figure}[h!]
\centering
\includegraphics[width=6cm]{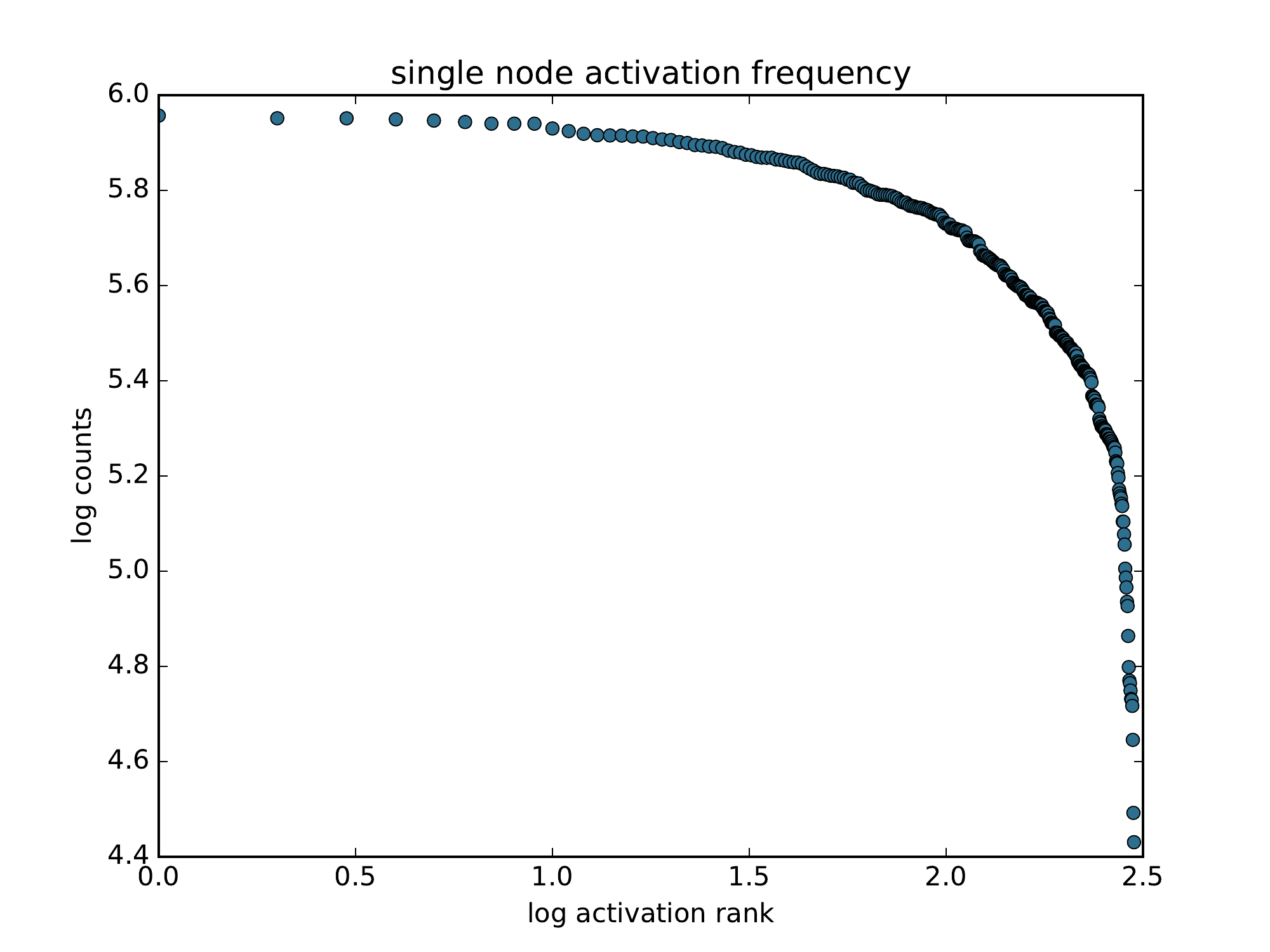}
\caption{Feed-forward net: Log-log plot of single node activation frequencies by rank}
\label{fig:ff_node_act_500_400_300_200_200_node_act_plot_layer_2}

\includegraphics[width=6cm]{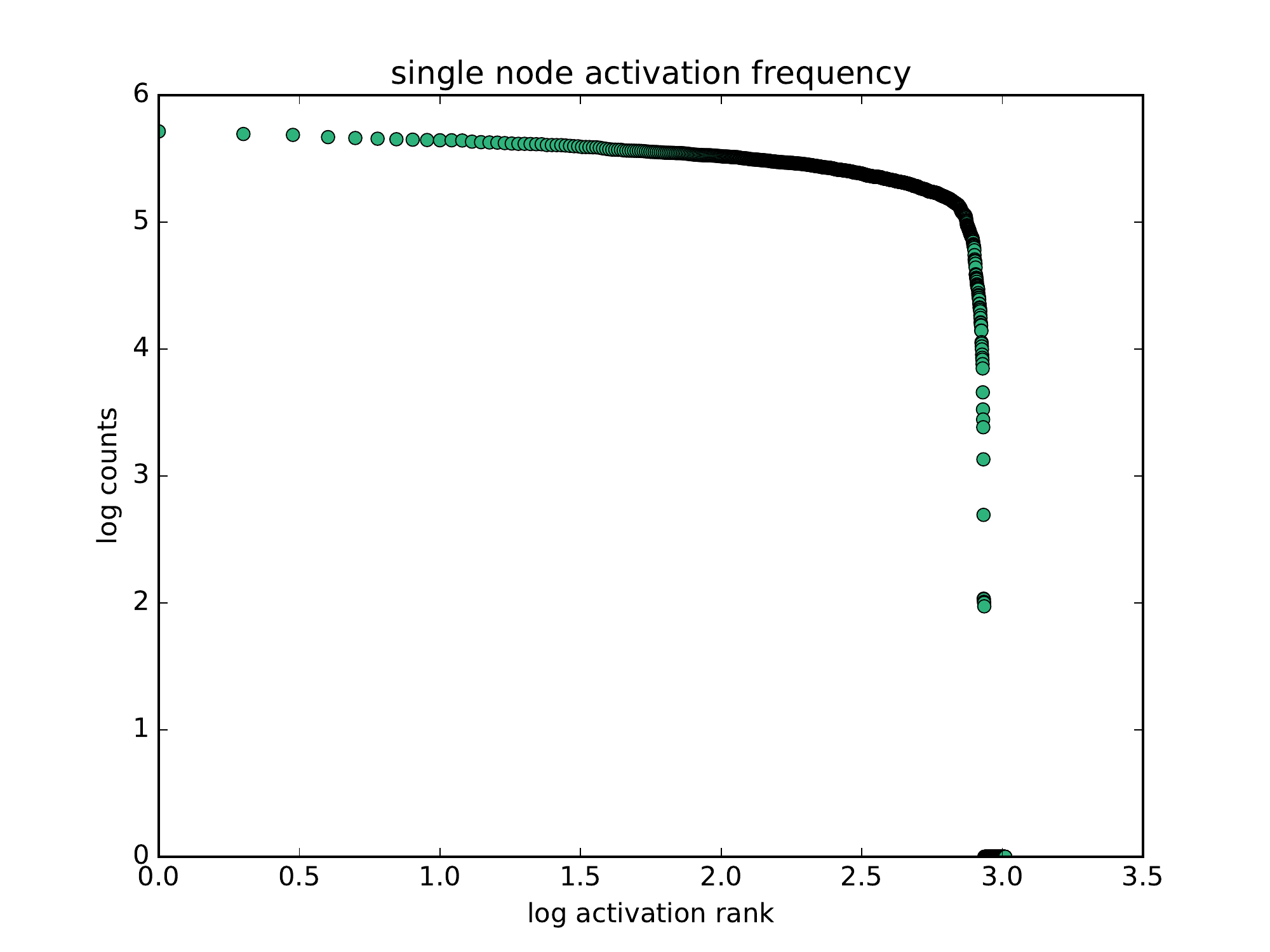}
\caption{CNN: Log-log plot of single node activation frequencies by rank}
\label{fig:cnn_node_act_conv3_local3_node_act_plot_layer_2}
\end{figure}

Figures \ref{fig:ff_node_act_500_400_300_200_200_node_act_plot_layer_2} and \ref{fig:cnn_node_act_conv3_local3_node_act_plot_layer_2} depict the behavior of feed-forward and CN network. The flat, nearly horizontal line in the latter architecture is again a sign of missing exponent whatsoever. 

As a last measure we employed the sum of activations defined as the average activations on each layer throughout the inference epochs. 

\begin{figure}[h!]
\centering
\includegraphics[width=6cm]{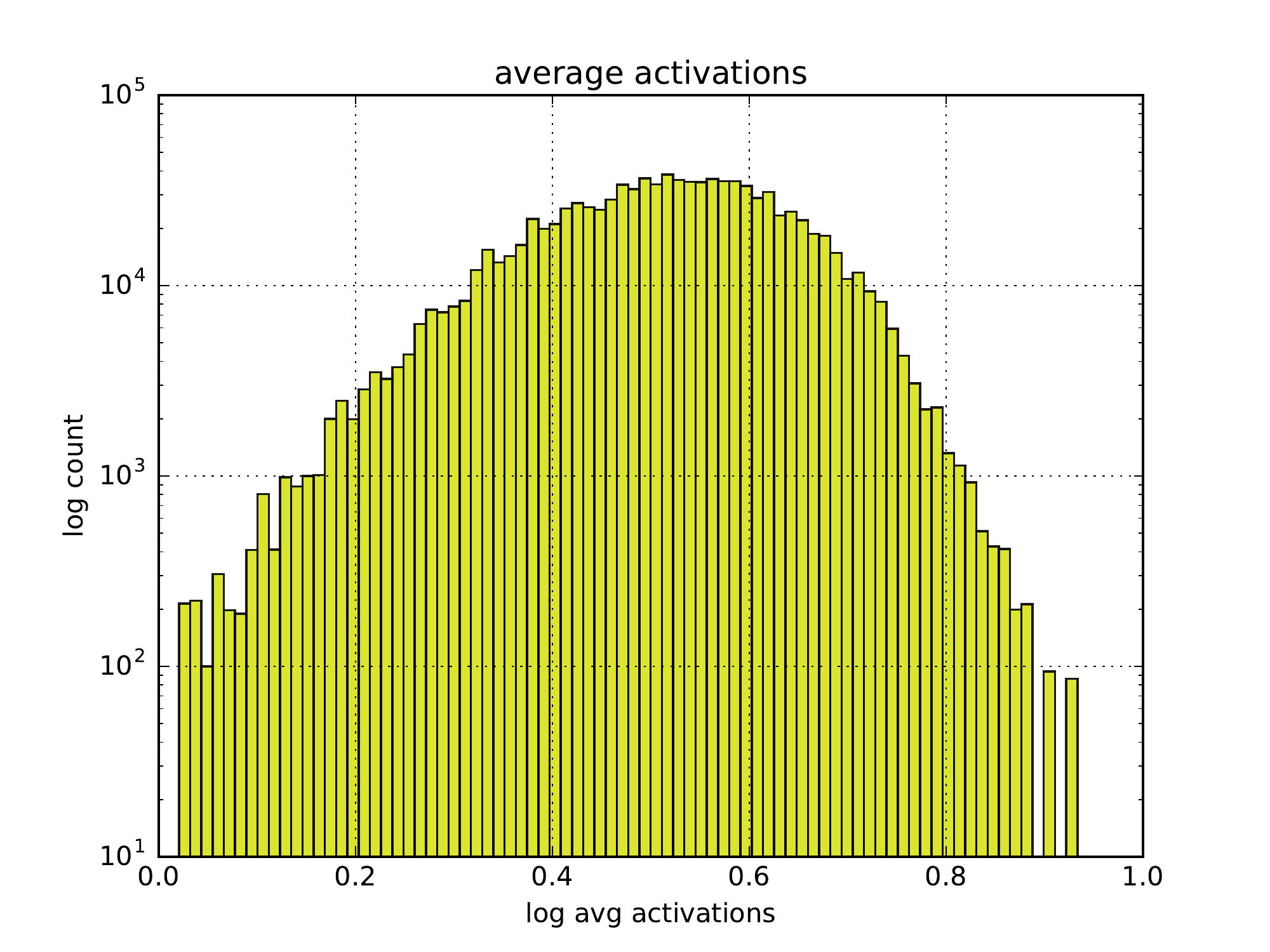}
\caption{Feed-forward net: Average layer activation distribution}
\label{fig:ff_sum_act_500_400_300_200_200_sum_act_plot_layer_4}

\includegraphics[width=6cm]{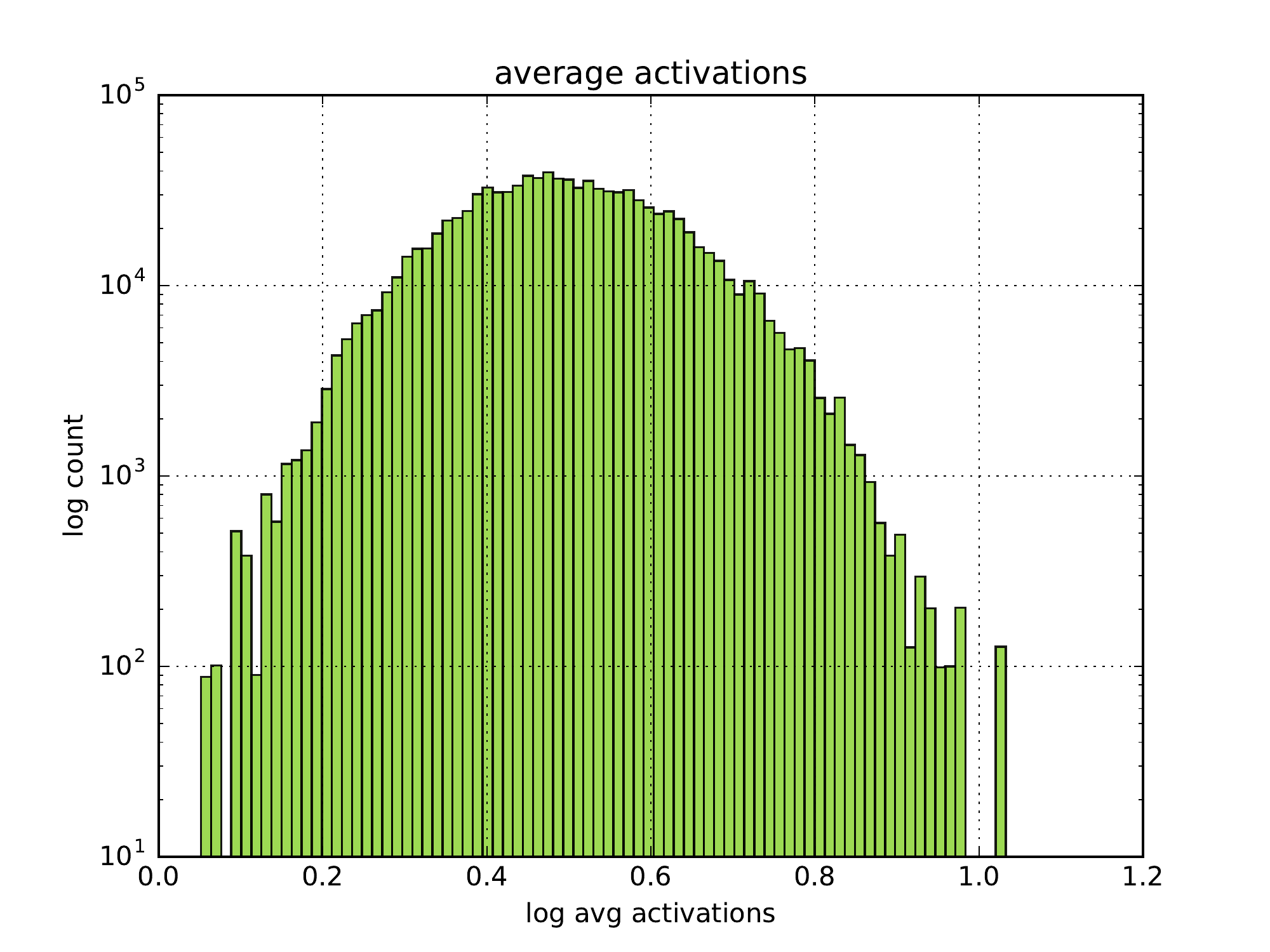}
\caption{CNN: Average layer activation distribution}
\label{fig:cnn_sum_act_conv3_local3_sum_act_plot_layer_4}
\end{figure}

Spontaneous and detectable criticality did not arise in classical architectures so the next step will be to create and experiment with systems that have induced criticality and learning rules that take into account criticality. Our first approach was to grow a fully connected net using the preferential attachment algorithm to induce at least some power law in node weights, and use the fully connected net as a hidden to hidden module. We further experimented with different solutions, regarding input and read out of activations from this hidden to hidden module, without changing the power law distribution. (This would simulate a system located at a critical state, with power law weight distribution). Our findings so far show that learning in these systems is very unstable without any advancement in learning and inference. The fundamental missing part is how to naturally induce a critical state in a network, which is equipped with learning rules that inherently take into account the critical state. For that we need new architectures and new learning rules, derived from the critical point equations (\ref{c_p}).

\section{Summary and outlook}
\label{sect:outlook}
Summary: In this article we make our first steps in investigating the relationship between criticality and deep learning networks. After a short introduction of criticality in statistical physics and real world networks we started with the theoretical setup of a fully connected layer. We used continuous mean field approximation techniques to tackle the partition function of the system ending up with a system of differential equations that determine the critical behaviour of the system. These equations can be the starting point for a possible network architecture with induced criticality and learning rules exploiting criticality. After that we presented results of experiments aiming to find traces of power law distributions in current deep learning networks such as multilayer feed-forward nets, convolutional networks and autoencoders. The results - except for the autoencoder - were affirmative in the negative sense, setting up as next the necessity to create networks with induced criticality and learning rules that exploit the critical state. 

Outlook: Obviously the fully connected layer, which can be solved analytically on the theoretical side is of limited importance, as it translates into a rather simplistic architecture; more realistic, widely used set-ups, e.g. convolutional or recurrent nets, do very well contain the feed-forward mechanism, but are strongly deviating  and hence only partially mapped to our theoretical treatment; it would definitely be essential to address theoretically the convolution mechanism of deep nets and establish a link between the theoretical and experimental side; also inducing criticality into the net via eq. (\ref{c_p}) could prove beneficial and might very well affect learning behavior and flow on the surface on the loss function.

\appendix

{\Large \bf  \centering{Appendix} \par}
\bottomtitlebar

\section{Different unit values}
\label{app1}
We here show that the partition function with Hamiltonian 
\begin{equation}
\label{H01}
\centering
H_{\{0,1\}} = \sum_{ij}a_iw_{ij}a_j + h_ia_i
\end{equation}
who's units are taking values in $\{0,1\}$ has the same qualities as encoded in the partition function with Hamiltonian $H_{\{\pm 1\}}$, who's units take values in $\{\pm 1\}$.

We rewrite the Hamiltonian in (\ref{H01}) with units taking values in $\{\pm 1\}$ (using Einstein's summation convention over double indices) :

\begin{equation}
\label{H01p}
\centering
H = \frac{1}{4}(1-u_i)w_{ij}(1-u_j) + \frac{1}{2}h_i(1-u_i) 
\end{equation}

where the $u_i$ and $v_i$ take values in $\{\pm 1\}$. Carrying now the multiplications in (\ref{H01p}) yields

\begin{align}
\label{H01m}
H &= \frac{1}{4}\sum_{ij}w_{ij}
+ \frac{1}{4}u_iw_{ij}u_j\nonumber \\
&- \frac{2}{4}u_iw_{ij}\mathbbm{1}_j 
- \frac{1}{2}h_iu_i + \frac{1}{2}h_i\\ \nonumber
&= \frac{1}{2}(c + u_iw_{ij}u_j 
+ h'_i u_i) \nonumber
\end{align}

with $h_i' = -w_{ij}\mathbbm{1}_j - h_i$. Hence when computing the partition Z with (\ref{H01p}) we obtain

\begin{equation}
\centering
Z = \sum_{u\in\{\pm 1\}}e^H = e^c\sum_{a\in\{0,1\}}
e^{ a_iw_{ij}a_j + h'_ia_i } 
\end{equation}

where the right hand side is the original Hamiltonian with a shifted coupling $h'$. The additional constant $c$ factors out completely and hence when taking the logarithm and the second derivative it won't change the outcome. Also we note that the second derivative wrt. $h'$ is $\partial_{h'h'} = \partial_{hh}$.

\section{First order contribution}
\label{app2}

We consider here the Hamiltonian of the bi-partite graph connected via  weight matrix $w$ (with Einstein summation convention):

\begin{equation}
\centering
H_b = u_iw_{ij}v_j
\end{equation}

with the free energy 

\begin{equation}
\centering
F_b = -\ln{\sum_{u,v\in\{\pm 1\}}\exp( u_iw_{ij}v_j )}
\end{equation}

Without any loss of generality we set the temperature $T=1$, and we won't keep track of it. 
Carrying the partial sum over $v_i$ yields

\begin{align}
F_b &= -\ln{\sum_u\prod_j\big[\exp(u_iw_{ij}) + \exp(-u_iw_{ij})\big]}\\
&=-\ln{\sum_u\prod_j\big[2\cosh(u_iw_{ij}) \big]}\nonumber
\end{align}

The sum over the $v_i$ is understood as a collection of $2^N$ terms, each corresponding to a unique combination of 0's and 1's in the vector of length $N$ representing that specific state of the spins; however, the sum can be conveniently written as a product of $N$ binary summands, where each contains exactly the two possible states of the $i$th spin - this is where the product over $j$ comes from in upper formula.
Expanding now to lowest order in $w$ we obtain

\begin{align}
F_b &\sim  -\ln{\sum_u\prod_j(1 + \frac{(u_iw_{ij})^2}{2} )}\\
&\sim -\ln\sum_u\exp{\sum_j\frac{(u_iw_{ij})^2}{2}}\nonumber\\
&=-\ln\sum_u e^{H(u_i)}\nonumber
\end{align}

where $H(u_i)$ is the Hamiltonian of the fully connected graph, defined as (Einstein summation convention) 

\begin{align}
\label{hopf_bolz}
H(u_i) &= \sum_j\frac{(u_iw_{ij})^2}{2}\\
&=\frac{1}{2}\sum_{ik}\underbrace{(\textstyle\sum_j w_{ij}w_{jk})}_{w'_{ik}}u_iu_k\nonumber\\
&=\frac{1}{2}\sum_{ik}u_iw'_{ik}u_k\nonumber
\end{align}

A few notes are in place regarding eq. (\ref{hopf_bolz}): the matrix $w'_{ik}$ is now symmetric by construction and hence mediates between equally sized (actually identical) layers; further, all higher terms of the $\cosh$ function are even, hence all contributions are higher order, symmetric interactions of the layer $u_i$ with itself.

\clearpage


\end{document}